\documentclass[lettersize,journal]{IEEEtran}
\usepackage{amsmath,amsfonts}
\usepackage{algorithmic}
\usepackage{algorithm}
\usepackage{array}
\usepackage[caption=false,font=normalsize,labelfont=sf,textfont=sf]{subfig}
\usepackage{textcomp}
\usepackage{stfloats}
\usepackage{url}
\usepackage{verbatim}
\usepackage{graphicx}
\usepackage{cite}
\usepackage{threeparttable}

\makeatletter
\renewcommand{\maketag@@@}[1]{\hbox{\m@th\normalsize\normalfont#1}}%
\makeatother

\usepackage{tikz,xcolor,hyperref}

\definecolor{lime}{HTML}{A6CE39}
\DeclareRobustCommand{\orcidicon}{
\begin{tikzpicture}
\draw[lime, fill=lime] (0,0)
circle[radius=0.16]
node[white]{{\fontfamily{qag}\selectfont \tiny \.{I}D}}; 
\end{tikzpicture}
\hspace{-2mm}
}
\foreach \x in {A, ..., Z}{%
\expandafter\xdef\csname orcid\x\endcsname{\noexpand\href{https://orcid.org/\csname orcidauthor\x\endcsname}{\noexpand\orcidicon}}
}

\begin{document}

\title{Coupled Modeling and Fusion Control for a Multi-modal Deformable Land-air Robot}

\author{Xinyu Zhang\hspace{-1.5mm}\orcidA{},~\IEEEmembership{Member,~IEEE}, Yuanhao Huang\hspace{-1.5mm}\orcidB{},~\IEEEmembership{Graduate Student Member,~IEEE}, Kangyao Huang\hspace{-1.5mm}\orcidC{}, Ziqi Zhao, Jingwei Li, Huaping Liu\hspace{-1.5mm}\orcidD{},~\IEEEmembership{Senior Member,~IEEE}, and Jun Li

\thanks{Xinyu Zhang, Ziqi Zhao, Jingwei Li, Jun Li are with the School of Vehicle and Mobility, Tsinghua University, Beijing, P.R.China (e-mail: xyzhang@tsinghua.edu.cn; zhjtuzi5782@gmail.com; lijw00123@163.com; lj19580324@126.com)}

\thanks{Yuanhao Huang is with the School of Vehicle and Mobility, Tsinghua University, Beijing, P.R.China, and also with the School of Aviation, Inner Mongolia University of Technology, Hohhot, Inner Mongolia, P.R.China (e-mail: huangyuanhao\_work@163.com)}%

\thanks{Kangyao Huang and Huaping Liu are with the Department of Computer Science and Technology, Tsinghua University, Beijing, P.R.China (e-mail: kangyao.huang@outlook.com; hpliu@tsinghua.edu.cn)}}

\markboth{Journal of \LaTeX\ Class Files,~Vol.~14, No.~8, August~2021}%
{Shell \MakeLowercase{\textit{et al.}}: A Sample Article Using IEEEtran.cls for IEEE Journals}


\maketitle

\begin{abstract}

A deformable land-air robot is introduced with excellent driving and flying capabilities, offering a smooth switching mechanism between the two modes. An elaborate coupled dynamics model is established for the robot, including rotors, chassis, suspension, and the deformable structure. In addition, a model-based controller is designed for landing and mode switching in various unstructured conditions, such as slopes and curved surface. And considering locomotion and complex near-ground situations to achieve cooperation between the two fused modalities. This system was simulated in ADAMS/Simulink and a tested with hardware-in-the-loop system was constructed for testing in various slopes. With a designed controller, the results showed the robot is capable of fast and smooth land-air switching, with a 24.6 \% faster landing on slopes. The controller can also reduce landing offset and impact force more effectively than the normal control method at 32.7 \% and 34.3 \%, respectively.

\end{abstract}


\section{Introduction}

\IEEEPARstart{R}{obots} are commonly used in search and rescue operations due to the safety they provide. Aerial robots offer stable, fast, and reliable remote sensing information at high altitudes due to their convenience and flexibility \cite{martinez2022aerial,zhao2022offshore}. While single-locomotion robots often struggle to adapt in highly variable or uncertain environments, land-air robots offer both flying and driving capabilities for increased spatial flexibility, thereby reducing the energy demands of continuous flight.

Hybrid terrestrial and aerial quadrotors, generally composed of a quadrotor hinged at the center of a cylindrical cage or a pair of coaxial passive wheels, are a common dual-motion robot type \cite{Fan2019,Kalantari2014a,Dudley2015a}. However, in this design, terrestrial motion depends on aerodynamic power generated by the rotors, which generates downwash that can negatively affect driving performance \cite{wen2019numerical}. Another class of robots exhibits both active aerial and terrestrial motion capabilities, which requires a driving or walking base \cite{tan2021multimodal}. For instance, a robot may utilize suction cups attached to its legs to probe the surface of a bridge during inspection and maintenance \cite{Ratsamee2016a}. The Drivocopter developed by Jet Propulsion Laboratories uses four oblate spherically-shaped elastic cages as wheels, thus protecting the propellers and absorbing shock forces during landing \cite{Kalantari2020a}. Chassis are also widely used in multi-modal robot designs to enhance ground motion performance \cite{Tan2021b,Hu2018a,Sharif2019c}.

Deformable robots can adjust their structures to adapt in different environments. Rotorcrafts typically require larger propeller blade areas for increased lifting force, which can be produced by increasing the number of rotors or increasing the propeller size. However, driving on the ground benefits from a smaller and more flexible base for enhanced maneuverability, especially when passing through narrow or unstructured environments. As such, a spatial utilization conflict often occurs in multi-modal systems. To address this issue, reusable and deformable structures have been proposed in recent years. In addition to sharing actuators, the flying star can adjust its body height, using a servo motor, while passing through tunnels \cite{Meiri2019a}. A robot developed by Kossett et al. was able to shrink its coaxial rotors adherent to its body, thereby overcoming most obstacles \cite{Kossett2010a}. A similar concept has been used in other studies \cite{Wang2019b}, where propeller shielding was modified to a wheel-shaped structure. This allowed the robot to transform between terrestrial and aerial locomotion using servos. However, limited by redundant mechanisms and actuators, robots still have problems such as insufficient load capacity and weak operating ability. At the same time, researchers need to give more thought to maintaining a reasonable balance between flight and ground driving abilities, which is the unique contribution of the design of the robot. 

\begin{figure}[t]
\centering
\includegraphics[width=6.5cm]{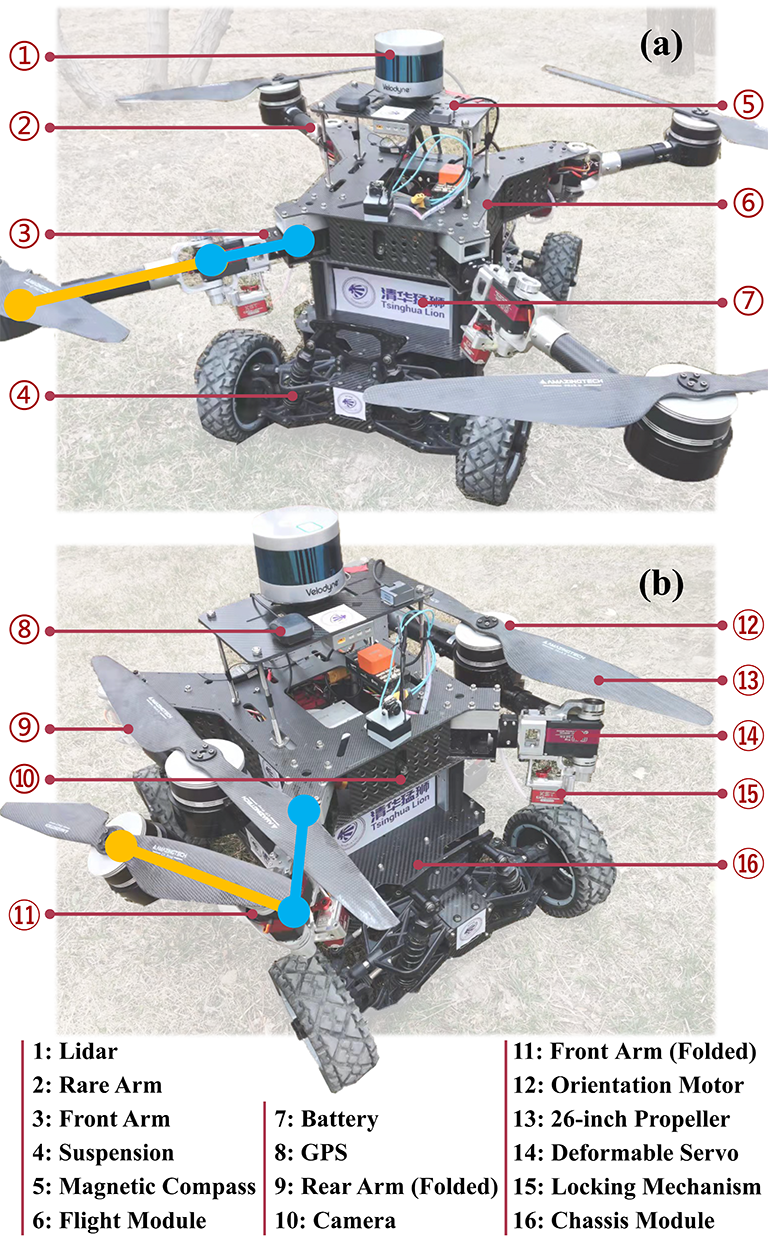}%
\caption{The structural design of the proposed robot. (a) With the arms deployed, the robot generates a large thrust force and load capacity with a 26-inch propeller. (b) Driving flexibility and passage capacity are then improved with the arms folded.}
\label{structure}
\end{figure}

In addition, the proposed robot, which can be considered a coupled aircraft and chassis, must exhibit stability and smoothness during mode switching. Proportional–integral–derivative (PID) \cite{ramon2019autonomous}, nonlinear model predictive control (NMPC) \cite{zhang2021robust}, reinforcement learning (RL) \cite{lee2018vision}, and other controller methods have been used to provide stable single-mode control. The trajectory can also be planned using EGO-planner\cite{zhou2020ego} or jerk-limited trajectory (JLT) \cite{paris2020dynamic}. External disturbances and primary actuators typically vary with the environment. The impact of the ground can also influence the robot during mode switching, which presents significant challenges for switching control in land-air robots \cite{zhang2022intelligent}.

In a previous study, we developed a multi-modal robot with a hexacopter and focused on continuous landing control using an MPC controller\cite{tan2021multimodal}. On this basis, a deformable land-air prototype was introduced in \cite{zhang2022multi}, the detailed structure of which is shown in Fig. \ref{structure}. This paper further introduced robot design details, including mechatronics, modeling, and control. Specifically, a controller was designed for mode-switching tasks by considering slopes and ground effect disturbances. The contributions of this study can be summarized as follows:

\begin{itemize}
    \item We introduce an innovative mechatronics design for variable-structure land-air robots, which taking into account the balance of flight and ground movement, including folding arms, locking mechanisms, and a fixed-propeller motor.
    \item A coupled dynamics model is proposed to describe the multi-modal motion of land-air robots. The deformable structure is also analyzed dynamically to access its states.
    \item Robot dynamics and ground effect models are combined and a fusion controller based on Linear–quadratic regulator (LQR) and JLT is proposed for continuous take-off and landing tasks in sloped scenes. The results of both simulations and experiments provided strong evidence for the effectiveness and robustness of this configuration.
\end{itemize}

The remainder of this paper is organized as follows. Section II describes the specific robot design, including novel structures and mechatronics. Section III provides details for the kinematic and dynamics models. Section IV presents the results of validation experiments and simulations, while Section V concludes the study.

\section{Mechatronics Design}

Robot design depends on operating context. As such, the quadrotor cannot simply be combined with the chassis, as robot capabilities must be balanced in multi-modal scenarios. For the proposed robot, energy system analysis, power system selection, and a lightweight design improved robot duration and operating times. The sharing of energy and control systems also helped to reduce the robot's weight, while a deformable structure ensured sufficient flight power with flexible ground motion. In this section, we primarily introduce and analyze the robot design, deformable structure, and mechatronics. The key parameters shown in Table \ref{specification} were used in high-fidelity model construction.

\begin{table}[h]
\renewcommand\arraystretch{1.1}
\begin{center}
\vspace{-3 mm} 
\caption{Robot specifications.}
\label{specification}
\begin{tabular}{c c c}
\hline \text {  \textbf{Height} } & \text { Body } & 552 {~mm} \\
\hline \text { \textbf{Weight} } & \text { Total (excluding payload) } & 20.62 {~kg} \\
& \text { Battery } & 5.74 {~kg} \\
& \text { Flight Module } & 7.81 {~kg} \\
& \text { Chassis } & 7.07 {~kg} \\
\hline \text { \textbf{Speed} } & \text { Max Flying Speed } & 7.47 {~m}/{s} \\
& \text { Max Driving Speed } & 15.10 {~m}/{s} \\
\hline \text { \textbf{Flight Module} } & \text { Flight Motor } & 5 {Nm} \\
& \text { Kv Value } & 170 \\
& \text { Propeller } & 26 {inches} \\
& \text { Steering Gear} & 8 {Nm} \\
\hline \text { \textbf{Chassis} } & \text { Kv Value } & 800 \\
& \text { Driving Motor } & 12 {Nm} \\
\hline
\end{tabular}
\end{center}
\end{table}

\begin{figure}[t]
    \centering
    \includegraphics[width=6cm]{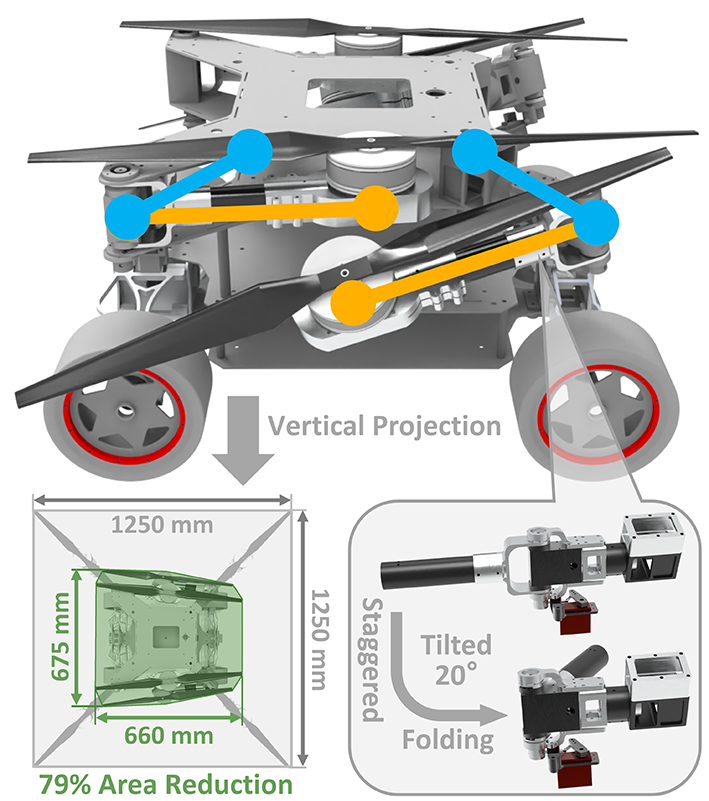}
    \caption{Self-deforming structures provide an innovative approach to balancing the flying and driving capabilities of land-air robots. Here, the blue lines indicate fixed arms and the yellow lines represent folding arms. The staggered folding method retains larger propellers while minimizing robot size to ensure stable movements in driving mode.}
    \label{deformable structure}
\end{figure}

The proposed power system uses 26-inch propellers to generate the required thrust. However, in flying mode, the robot assumes a standard quadrotor configuration with a maximum length and width of 1250 mm. These dimensions make it difficult to pass through complex unstructured scenes, both outdoors and indoors. In addition, the scattered mass distribution reduces dynamic stability and can induce vibrations in the robot \cite{saab2018robotic}. As such, a deformable structure is required to reduce robot size when moving on the ground. Fig. \ref{deformable structure} demonstrates this staggered folding strategy, in which the platform enters driving mode as the steering gear drives the four arms to rotate independently. The front arms are then tilted by 20 degrees while the rear arms remain horizontal. This design avoids arm collisions and further reduces the robot's size. As a result, the robot's vertical projection area is reduced by 79\% compared with flying mode. The robot, limited to a span of 700 mm, can then easily navigate through narrow passages or gates. Furthermore, the deformable arms are directly driven by servos and a locking mechanism is included as the last component of flying safety.

\begin{figure}[t]
    \centering
    \includegraphics[width=7cm]{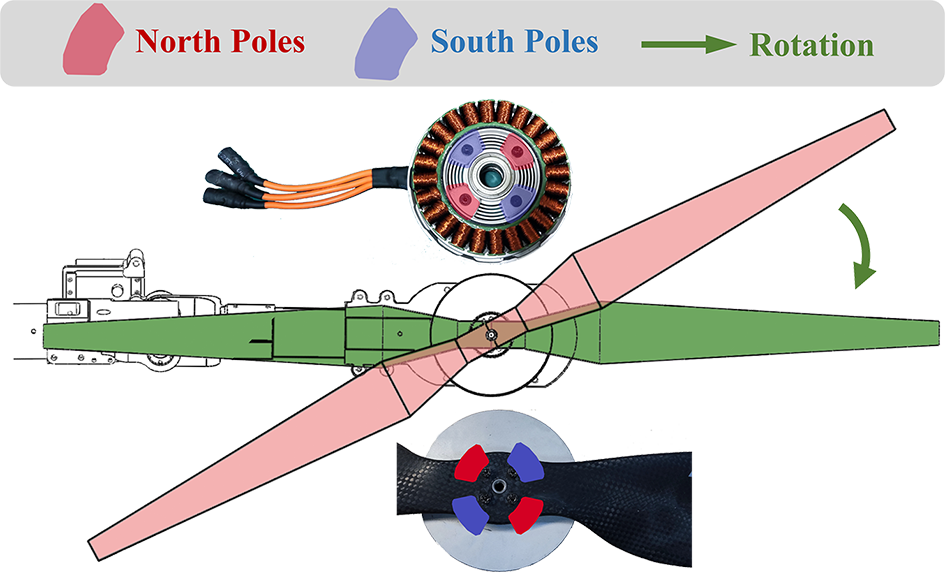}
    \caption{The motor limits propeller orientation using the positions of non-contact magnets. After detecting a collision or artificially moving the propeller, position can be quickly restored in 1-2 s.}
    \label{orientation motor}
\end{figure}

\begin{figure}[b]
    \centering
    \includegraphics[width=8.6cm]{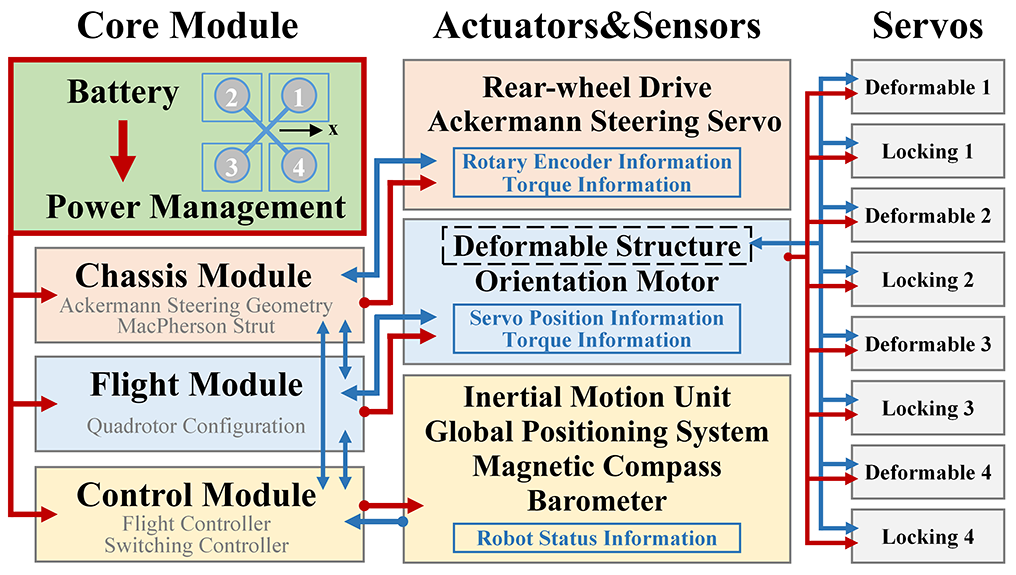}
    \caption{An overview of the hardware system. The red lines represent the power supply and the blue lines indicate signal communication. The flight and chassis modules share the power supply and controller. The serial number of each arm is also shown in the figure. When transitioning from flight to driving mode, the robot must fold the rear arms first and then the front arms. Otherwise, the front arm is unfolded first, followed by the rear arm.}
    \label{hardware}
    \vspace{-5 mm} 
\end{figure}

When driving on land, rotors should maintain propellers and arms in the same orientation to avoid collisions between the propeller and the body (see Fig. \ref{orientation motor}). Thus, relying on mechanical or electrical stops may increase system complexity and a magnetic mechanism was thus introduced to maintain propeller position. This magnet can generate attractive forces without contact and was used for fixation and restraint \cite{moreno2018automated}. In this study, four magnets with different polarities were staggered on the inner and outer walls of the motor as a passive positioning mechanism. The magnets were then distributed vertically with alternating polarity (avoiding contact), which reduced power loss in the motor. In this way, the magnets could hold the propellers in a static state and quickly return to a favorable position in the case of rotation or offset.

A mode switching controller was used to operate the auto-folding and auto-locking mechanisms. The flight module included 8 servos for driving and locking the deformable arms. Four of these servos were dual-axis steering gears used for automated degree of freedom management during folding. The other four were locking servos used to insert fixed pins into the locking pin. The servo controller was equipped with a potentiometer array for fine-tuning of eight servo positions, to reduce the gap caused by fit tolerance, as shown in Fig. \ref{hardware}. This configuration effectively reduced calculation pressure during flight control, which provided increased safety.

\section{Kinematic and Dynamics Models}

\subsection{Kinematic Model}

\begin{figure}[t]
    \centering
    \includegraphics[width=8cm]{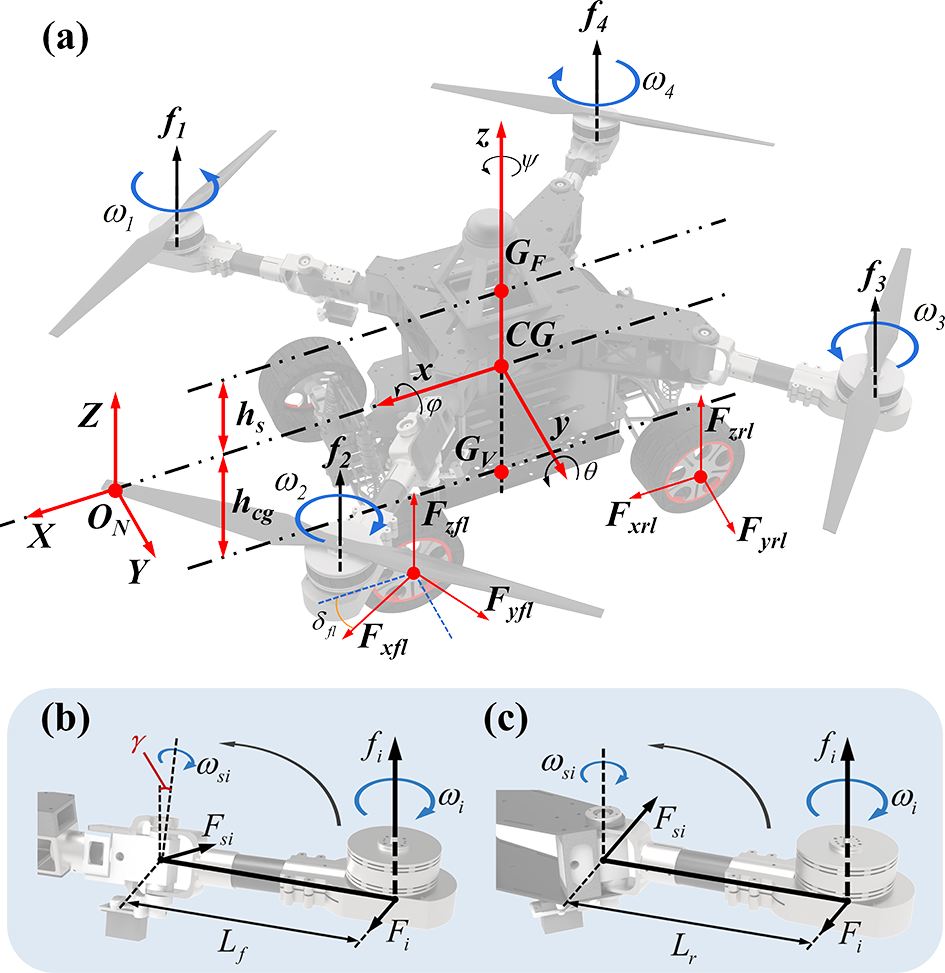}
    \caption{An equivalent mathematical model for the land-air robot. (a) Modeling of the overall system and flight module. (b), (c) Modeling of the folding component on the front and rear arms.}
    \label{math model}
\end{figure}

The land-air robot included a chassis module with a fixed connection to the flight module. Motors then provided force and torque, while the chassis with rear-wheel drivers and a Macpherson strut provided ground propulsion. This land-air platform constitutes a strongly coupled system with multiple degrees of freedom. Variable pose relationships in the model were used to develop a control algorithm for motion and mode switching, as shown in Fig. \ref{math model}. Several assumptions were made while considering the moment of inertia and fusion control during movement:

\begin{itemize}
    \item Connections between the frame of the rotor mechanism and the chassis were rigid and symmetric.
    \item The four propellers were rigid and experienced no blade flapping.
    \item The arms were rigid during flight and surface movements.
\end{itemize}

The relative pose relationship of the land-air platform was represented by a body coordinate system $(C G-x y z)$ and a global coordinate system $\left(O_{N}-X Y Z\right)$. These were used to establish additional coordinate systems, the body angular velocity in the body coordinate
system $\omega_{n}=\left[\begin{array}{lll}
\phi & \theta & \psi
\end{array}\right]^{T}$ 
, and angular velocity in global coordinate system 
$\omega_{b}=\left[\begin{array}{lll}
p & q & r
\end{array}\right]^{T}$. Here, $\phi$, $\theta$, and $\psi$ represent the roll, pitch, and yaw angles for the body relative to the global coordinate system. The terms $p$, $q$, and $r$ denote the roll, pitch, and yaw angular velocities for the body relative to the body coordinates, respectively. In this part, the position of the body's center of gravity in the global coordinate system is defined as $\overrightarrow{r_{n}}$, the acceleration of the land-air platform in the global coordinate system is denoted $\vec{F}=m \ddot{\vec{r}}$, the angular velocity of the body in the body coordinate system is represented by $\overrightarrow{\omega_{b}}\left[\begin{array}{lll} \dot{p} & \dot{q} & \dot{r} \end{array}\right]^{T}$. As in \cite{chen2016robust}, the angular velocity in the global coordinate system is given by $\overrightarrow{\omega_{n}}\left[\begin{array}{lll} \dot{\phi} & \dot{\theta} & \dot{\psi} \end{array}\right]^{T}$. And the transformation matrix from the body coordinate system to the global coordinate system can then be expressed as ${ }^{n} \mathrm{T}_{b}$. 

\begin{equation}
{ }^{n} \mathrm{T}_{b}=\left[\begin{array}{ccc}
C_{\psi} C_{\theta} & C_{\psi} S_{\theta} S_{\phi}-S_{\psi} C_{\phi} & C_{\phi} C_{\psi} S_{\theta}-S_{\phi} S_{\psi} \\
S_{\psi} C_{\theta} & S_{\psi} S_{\theta} S_{\phi}+C_{\psi} C_{\phi} & S_{\psi} S_{\theta} C_{\phi}-C_{\psi} S_{\phi} \\
-S_{\theta} & C_{\theta} S_{\phi} & C_{\theta} C_{\phi}
\end{array}\right]
 \end{equation}

The force defined in the body coordinate system $(C G-x y z)$ can also be transformed into the global coordinate system $\left(O_{N}-X Y Z\right)$ using a transformation matrix ${ }^{n} \mathrm{T}_{b}$.

The transformation matrix for the angular velocity ${ }^{n} \mathrm{D}_{b}$, from body coordinates to global coordinates, can be expressed as:
\begin{equation}
{ }^{n} \mathrm{D}_{b}=\left[\begin{array}{ccc}
1 & S_{\phi} T_{\theta} & C_{\phi} T_{\theta} \\
0 & C_{\phi} & -S_{\phi} \\
0 & S_{\phi} / C_{\theta} & C_{\phi} / C_{\theta}
\end{array}\right],
\end{equation}
where $C$ represents $cos$, $S$ denotes $sin$, and $T$ denotes $tan$. The rotation equation for the flight module in the body coordinate system is given by:
\begin{equation}
J_{b} \cdot \frac{d \overrightarrow{\omega_{b}}}{d t}=\overrightarrow{M_{b}}-\overrightarrow{\omega_{b}} \times J_{b} \cdot \overrightarrow{\omega_{b}},
\label{rotation}
\end{equation}
where $\overrightarrow{M_{b}}$ is the total torque in the system, $J_{b}$ is the rotational inertia matrix for the flight module in the body coordinate system, $M_{g}$ is the torque generated by the gyroscopic effect, and $M_{d}$ is the torque generated by aerodynamic friction. The magnitudes of these terms are estimated in subsequent sections. The fixed coordinate system for the flight module body transfers angular acceleration into the global coordinate system, which can be estimated using the transformation matrix ${ }^{n} \mathrm{D}_{b}$. This model can be simplified by assuming the inertia matrix $J_{b}$ is a diagonal matrix as follows:
\begin{equation}
J_{b}=diag[J_{x}, J_{y}, J_{z}].
\end{equation}
Two sub-models are analyzed in the following subsections and used to calculate forces and moments acting on the flight module, to conduct further analysis.

\subsection{Flight Module Dynamic Model}

This section establishes an equivalent dynamic model for the proposed flight module. The quadrotor was treated as a coupled rigid body with six degrees of freedom, as shown in Fig. \ref{math model}(a). The Newton-Euler equation suggests a dynamic model of the quadrotor aircraft can be expressed as:
\begin{equation}
M_{b}=J_{b} \dot{\omega}_{b}+\omega_{b} \times J_{b} \omega_{b}+M_{g}+M_{d}.
\end{equation}
A synthesis of gyroscopic torque $M_{g}$ and aerodynamic friction torque $M_{d}$ can then be represented by:
\begin{equation}
M_{g}=\sum_{i=1}^{4} \overrightarrow{\omega_{b}} \times J_{b}\left[0 \quad 0(-1)^{i+1} \Omega_{i}\right]^{T},
\end{equation}
\begin{equation}
M_{d}=\operatorname{diag}\left(d_{\phi}, d_{\theta}, d_{\psi}\right) \dot{\zeta},
\end{equation}
where $J_{b}=\operatorname{diag}\left(d_{\phi}, d_{\theta}, d_{\psi}\right)$ is the moment of inertia for each rotor and $d_{\phi}$, $d_{\theta}$, and $d_{\psi}$ are the corresponding aerodynamic damping coefficients. The translation equation for the quadrotor aircraft in a global coordinate system can then be derived as follows:
\begin{equation}
m \ddot{P}={ }^{n} D_{b} \cdot F+\left[\begin{array}{c}
0 \\
0 \\
-m g
\end{array}\right]-\left[\begin{array}{c}
d_{x} \dot{x} \\
d_{y} \dot{y} \\
d_{z} \dot{z}
\end{array}\right],
\end{equation}
where $d_{x}$, $d_{y}$, and $d_{z}$ respectively represent the introduced aerodynamic friction coefficients, used to calculate resistance to translational motion, and $F$ is the lift generated by the rotor.

Complex deformable arms can affect the payload and reliability of a land-air robot, as each rotor is fixed to the main body by an independent arm. The steering mechanisms in each arm enable the rotor to rotate independently about the main body. In the dynamic modelling process, the arms are approximated as rectangular cuboids \cite{falanga2018foldable} and the propeller generates yaw force during rotation, relying on force generated by the steering gear to maintain the fuselage structure shown in Figs. \ref{math model}(b) and (c). The moment of inertia for the arm can be expressed as:
\begin{equation}
\begin{array}{l}
J_{f i}=\frac{m_{i}}{12} L_{f i}^{2} \cdot  \cos (\gamma) \vspace{2ex} \cdot \left(w_{si}^{2}+L_{f}^{2}\right),\\
J_{r i}=\frac{m_{i}}{12} L_{r i}^{2} \cdot \left(w_{si}^{2}+L_{f}^{2}\right),
\end{array}
\end{equation}
where $m_{i}$ is the mass of the arm and $J_{i}$ is the moment of inertia for the arm around the axis of the steering gear. When the propeller is stationary, the only force ($F_{si}$) on the front arm deformation actuator is the component of gravity in the direction of the rotational plane. As the propeller rotates, this force is the result of the yaw moment $F_{i}$ and gravity in the radial direction of the rotating shaft. A dynamic equation for the deformable arm is then given by:
\begin{equation}
\operatorname{diag}\left(m_{i}\right)\left[\begin{array}{c}
\ddot{x} \\
\ddot{y} \\
\ddot{z}
\end{array}\right]=\left[\begin{array}{c}
F_{x i} \\
F_{y i} \\
F_{z i}
\end{array}\right],
\end{equation}
where $F_{x i}, F_{y i}, \text { and } F_{z i} (i=1 \ldots 4)$ are decoupling forces along the x-axis, the y-axis, and the rotating z-shaft of each motor, respectively. 

In this process, the rotation matrix decouples the torque and the reference coordinate is a global coordinate system. As the arm and the rotor rotate relative to the land-air platform, the inertia tensor for the arm does not rotate. This inertia tensor can be expressed as:
\begin{equation}
J_{a r m, i}=R_{z}\left(\theta_{i}\right) J_{a r m} R_{z}\left(\theta_{i}\right)^{T},
\end{equation}
where $R_{z}$ is the rotation matrix around the z-axis $(\theta_{i})$ in the global coordinate system and $i=(fl,fr,rl,rr)$. The moment of inertia for the motor does not change as it rotates around the z-axis. As such, the moment of inertia for the motor and rotor can be ignored and the steering gear joint can be considered a rigid body during flight and ground movement. Eq. (\ref{rotation}) suggests the rotation equation for the aircraft dynamics model can be represented in the body coordinate system as follows:
\begin{equation}
\left[\begin{array}{c}
\dot{p} \\
\dot{q} \\
\dot{r}
\end{array}\right]=\left[\begin{array}{l}
{\left[-d_{x} \cdot l \cdot p+F_{x}+q r\left(J_{y}-J_{z}\right)\right] / J_{x}} \\
{\left[-d_{y} \cdot l \cdot q+F_{y}+p r\left(J_{z}-J_{x}\right)\right] / J_{y}} \\
{\left[-d_{z} \cdot l \cdot r+F_{z}+p q\left(J_{x}-J_{y}\right)\right] / J_{z}}
\end{array}\right],
\end{equation}
where $J_{x}$, $J_{y}$, and $J_{z}$ are the moments of inertia around the x-axis, y-axis, and z-axis in the corresponding directions, respectively, and $F_{x}$, $F_{y}$, and $F_{z}$ denote actuator forces of each direction in the body coordinate system. The relationship between the force and the torque can be expressed as:
\begin{equation}
\left[\begin{array}{c}
F \quad F_{x} \quad F_{y} \quad F_{z}
\end{array}\right]^T=c_{\Omega} \cdot M_{4} \cdot\sum_{i=1}^{4} \omega_{i}^{T},
\end{equation}
where $c_{\Omega}$ is the thrust coefficient, $\alpha=45^{\circ }$ is half the angle between the arms of the quadrotor, $\omega_{i}(i=1…4)$ is the rotational speed of the rotor, and $M_{4}$ is the control matrix for the flight module.

\subsection{Chassis Dynamic Model}

\begin{figure}[t]
    \centering
    \includegraphics[width=7cm]{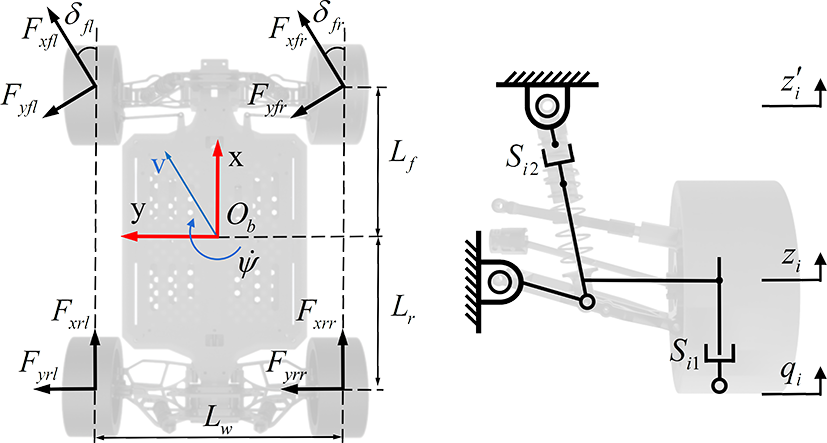}
    \caption{Mathematical modeling of an Ackerman chassis and a MacPherson suspension.}
    \label{chassis}
\end{figure}

In this study, the magic formula \cite{pacejka1992magic} was used to describe the tires, which were subjected to longitudinal forces to maintain speed. The lateral force was then approximated as follows:
\begin{footnotesize}
\begin{equation}{ll}
F_{y i}=D_{L} \cdot \sin \left[C_{L} \cdot \arctan \left\{B_{L} \alpha_{i}-E_{L}\left(B_{L} \alpha_{i}-\arctan \left(B_{L} \alpha_{i}\right)\right)\right\}\right],
\end{equation}
\end{footnotesize}
where $i=(fl,fr,rl,rr)$, $B_{L}$, $C_{L}$, $D_{L}$, and $E_{L}$ are parameters acquired by model fitting, and the longitudinal load $F_{Ni}$ can be expressed as:
\begin{equation}
F_{N i}=m_{i} \ddot{z}_{i}+c_{i}\left(\dot{z}_{i}-\dot{z}_{i}^{\prime}\right)+k_{i}\left(z_{i}-z_{i}^{\prime}\right),
\end{equation}
where $\dot{z}_{i}^{\prime}$ represents the sprung mass vertical displacement of a spring on the end of the wheel, $z_{i}$ represents the unsprung mass vertical displacement. As such, the y-direction component of the wheel in world coordinates system $F_{y i}$ is constrained by
\begin{equation}
F_{y i} \leq \sqrt{\left(\mu F_{N i}\right)^{2}-F_{x i}^{2}}.
\end{equation}
The results of wheel modelling were then combined with the chassis model. Fig. \ref{chassis} shows the dynamic response of the chassis structure for a given force and moment. The translation equation for the chassis (in the x and y directions) and the rotation equation for the yaw angle can be represented as follows:
\begin{equation}
\begin{cases}{l}
M_{b} a_{x}=\sum_{i} F_{L i}-\frac{1}{2} \rho c_{\omega} A v_{x}^{2}\vspace{0.5ex}, \\
M_{b} a_{y}=\sum_{i} F_{Q i}\vspace{0.5ex}, \\
J_{z} \ddot{\psi}=\frac{b}{2}\left(F_{L f r r}+F_{L r r}-F_{L f l}-F_{L r l}\right)+\\
\left(F_{Q f l}+F_{Q f r}\right) \cdot L_{f}-\left(F_{Q r l}+F_{Q r r}\right) \cdot L_{r}+U_{4},
\end{cases}
\end{equation}
where
\begin{equation}
\begin{array}{cc}
F_{L i}=F_{x i} \cos \left(\delta_{i}\right)-F_{y i} \sin \left(\delta_{i}\right)\vspace{0.5ex},\\
F_{Q i}=F_{x i} \sin \left(\delta_{i}\right)-F_{y i} \cos \left(\delta_{i}\right).
\end{array}
\end{equation}

The chassis suspension system shown in Fig. \ref{chassis} primarily consists of a chassis (a sprung mass $\dot{m}_{i}^{\prime}$), a suspension swing arm, a shock absorber, and wheels (an unsprung mass $\dot{m}_{i}$). The term $\dot{z}_{i}^{\prime}$ represents the sprung mass displacement in the vertical direction, $\dot{z}_{i}$ is the unsprung mass displacement, and $q_{i}$ is the ground disturbance. The constant terms $S_{i1}$ and $S_{i2}$ represent the damping, while $B_{i1}$ and $B_{i2}$ and $K_{i1}$ and $K_{i2}$ denote the stiffness of the shock absorber and wheel, respectively. The chassis suspension equation can then be established as follows:
\begin{equation}
\begin{cases}
-K_{1 i}\left(\dot{z}_{i}^{\prime}-\dot{z}_{i}\right)-\left(B_{1 i}+B_{2 i}\right)\left(\dot{z}_{i}^{\prime}-\dot{z}_{i}\right)+f_{a}-f_{d}=m_{i} \ddot{z}_{i}^{\prime}, \\
K_{1 i}\left(z_{i}-q_{i}\right)+K_{2 i}\left(\dot{z}_{i}^{\prime}-\dot{z}_{i}\right)+\left(B_{1 i}+B_{2 i}\right)\left(\dot{z}_{i}^{\prime}-\dot{z}_{i}\right)\\
-f_{a}=m_{i} \ddot{z}_{i},
\end{cases}
\end{equation}
where $f_{a}$ and $f_{d}$ represent actuator and disturbance forces, respectively. At this point, the state space for the suspension model can be described as:
\begin{equation}
\dot{X} =A x+B_{1} f_{a}+B_{2} q_{i}+B_{3} f_{d},
\end{equation}
where ${q}_i$ represents road disturbance.

\section{Fusion Controller Design}
\begin{figure*}[t]
\centering
\includegraphics[width=18cm]{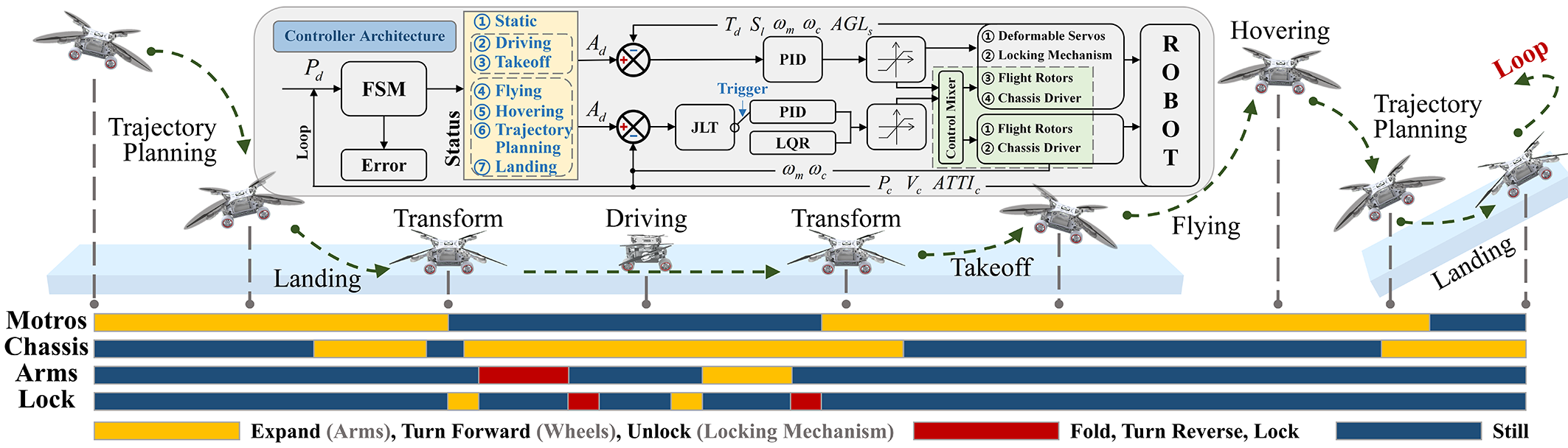}%
\caption{The finite state machine used to switch between motion states and modes. Each actuator exhibits a corresponding relationship with the motion state. The controller proposed in this paper was implemented using JLT and LQR.}
\label{controller}
\vspace{-2 mm}
\end{figure*}

Landing on slopes is a challenging task for land-air robots, especially with a heavy coupled chassis. This section describes an algorithm used to switch robot motion states, employing a finite state machine. We also consider takeoff and landing as optimization problems. In this process, fusion control during multi-modal motion is conducted using a JLT-based trajectory planning controller and an LQR-based auxiliary controller that considers the modeling of ground effects, as described below. The corresponding motion and controller systems are shown in Fig. \ref{controller}.

\subsection{Finite State Machine}
Both single-motion and multi-mode switching must be conducted in various situations. In these cases, the driving mechanism and control algorithm (used for multi-modal movements) also differ. The proposed design utilized a finite state machine (FSM) to achieve control switching during multi-modal robot motion. Basic movements and control logic in the deformable structure were incorporated into the FSM. A common scenario is described, which divides robot states into static, driving, transform, takeoff, flying, trajectory planning, and landing phases as follows:

\begin{itemize}
    \item $Static$: In this initial state, the robot is stable on the ground and its arms can be folded or unfolded. The robot then determines whether it has reached a preset point, engaging the FSM switch to $Driving$ or $Takeoff$ states.
    \item $Transform$: In this state, the robot arms can deform for conversion into different motion mode structures.
    \item $Drive$: The robot can drive on land in this state. If the arms are unfolded, the robot will unlock-fold-lock immediately during motion. The robot will then switch to a $Takeoff$ state when it reaches the takeoff point.
    \item $Takeoff$: Once the robot switches into this state, it will immediately unlock-unfold-lock the arms and turn the propellers for flight. The robot can take off vertically or with a run-up motion, switching to a $Flying$ state once it has left the ground by a distance of more than 500 mm (disturbances of ground effects can be negligible). 
    \item $Flying$: In this state, the robot flies to a set target point, after which the FSM automatically switches to a $Hovering$ state.
    \item $Hovering$: The robot maintains dynamic stability in this state. After reaching the hovering point, the robot can switch to a $Trajectory Planning$ state.
    \item $Trajectory$ $Planning$: The robot plans its trajectory in this state using the JLT-based controller discussed in Section IV.B. After approaching land, the robot automatically switches to a $Landing$ state.
    \item $Landing$: In addition to the trajectory plan, the robot is also influenced by ground effects when landing. As such, the robot relies on the LQR-based auxiliary controller, discussed in Section IV.C, for attitude stabilization. The robot lands vertically or with a run-up distance and can then switch to $Static$, $Driving$, or $Takeoff$ while executing a loop.
\end{itemize}

\subsection{Trajectory Planning}
In this section, modeling and optimization methods are proposed for takeoff and landing scenarios on sloped surfaces. Transitioning from air to a slope requires real-time planning for the specific environment and is limited by dynamic constraints of the robot. However, strongly coupled systems are generally safer and their controllers require fewer computations to achieve satisfactory real-time performance. In addition, takeoff and landing on a slope requires precise control of the robot's trajectory, attitude (specifically terminal posture), and speed. This study treated landing as an optimization problem for the position, velocity, and attitude, assuming each axis can be denoted as a triple integrator \cite{hehn2011quadrocopter}. The total optimization cost can be determined by independently solving these three spatial axes \cite{mueller2015computationally}.  Given a three differentiable motion vector $j(t)$, the jerk can be denoted as:
\begin{equation}
\boldsymbol{j_b}=\dddot{x}=\left(\dddot{x}_1, \dddot{x}_2, \dddot{x}_3\right).
\end{equation}
The mode switching trajectory generation is described as an optimization problem. The thrice differentiable robot trajectory at the final time $T$ can then be used to calculate the jerk cost function as follows:
\begin{equation}
min (J{\Sigma}) = \frac{1}{T} \int_{0}^{T}\|\boldsymbol{j}_b(t)\|^{2} d t, 
\label{cost}
\end{equation}
subject to
\begin{equation}
\begin{array}{l}
0 \leq f_{\min } \leq f \leq f_{\max }, \\
\|\omega\| \leq \omega_{\max }, \\
-v_{\max } \le  v(t) \le v_{\max }, \forall t \in\left[0, t_{\mathrm{end}}\right], \\
-j_{\max } \le j(t) \le j_{\max }, \forall t \in\left[0, t_{\mathrm{end}}\right].
\end{array}
\end{equation}

This cost function can be interpreted as an upper bound on the average of a product of the robot's control inputs. In this expression, $f$ describes the robot's flight thrust, which is limited by fixed-pitch propellers. The angular velocity is also limited to a sphere by the dynamic model or sensor saturation and can be expressed as a Euclidean norm $\|\omega\|$. The above parameters describe input constraints that work with the robot dynamics and motion constants used to limit the cost function. Furthermore, the robot is driven by two control inputs of the parameters $omg$ and $f$. In the controller, the nonlinear model and the generation of the trajectory are simplified. The cost function can then be decoupled into a per-axis cost $j_k$, which introduces position and velocity as follows:
\begin{equation}
\begin{array}{l}
J_{\Sigma}=\sum_{k=1}^{3}\frac{1}{T} \int_{0}^{T} j_{k(p, v)}(t)^{2} d t,
\end{array}
\end{equation}
where $p$ and $v$ are the position and velocity of the robot about each axis, respectively, $T$ is the preferred duration for the motive trajectory. Robot's state $\kappa$ can then be defined about the axes as:
\begin{equation}
\begin{array}{l}
\kappa = \left [ u\left ( t \right ), \dot{u}(t)\right ],
\end{array}
\end{equation}
where $\kappa$ is the expected state at the end moment $T$ of the trajectory, expressed as:
\begin{equation}
\begin{array}{l}
\kappa_{k} (T)=\hat{\kappa_{k}} , (k=1,\dots 6).
\label{expect}
\end{array}
\end{equation}

The desired position of the robot must be as accurate as possible during mode switching and the speed must be kept within a specific range to ensure safe operation. As such, the optimal trajectory can be determined using $\hat{\kappa}$. In the case of multiple trajectories, trajectory cost can be calculated using Eq. (\ref{cost}).

\subsection{Ground Effects and Motion Controller Design}
During robot motion, the lift and pitching moments may increase due to aerodynamic ground effects. This is particularly evident during takeoff and landing, especially on sloped surfaces, where pitching moment can increase by as much as 60\% \cite{paz2020cfd,kan2019analysis}. Since the robot cannot remain stably static with a tilted attitude, these ground effects may introduce instabilities and cause divergence of control. However, an equivalent ground effect model (EGEM) can be used to represent these effects by considering both altitude and forward velocity, wherein the velocity potential of a source can be represented as \cite{Cheeseman1955}:
\begin{equation}
\begin{array}{l}
\phi=-\displaystyle \frac{(R/4z)^{2}}{\sqrt{\left(x-x_{0}\right)^{2}+\left(y-y_{0}\right)^{2}+\left(z-z_{0}\right)^{2}}}.
\end{array}
\end{equation}
The ratio of thrust for ground effects and for air, from a single propeller, is given by:
\begin{equation}
\begin{array}{l}
\displaystyle \frac{T_{I G E}}{T_{O G E}} = \frac{1}{1-\frac{(R / 4 z)^{2}}{1+\left(|V| / v_{i}\right)^{2}}},
\end{array}
\end{equation}
where $T_{IGE}$ and $T_{OGE}$ denote the thrust with and without ground effects, respectively. In addition to these contributions, the controller should also consider a dynamics model for the wheel and suspension metrics, including an angular velocity $[\theta , \psi]^T$ and an angular acceleration $[\dot{\theta}, \dot{\psi}]^T$. The chassis dynamics model can then be linearized and converted to a state space as follows:
\begin{equation}
\begin{array}{l}
\dot{X}(t)=A X + B U+ G_{d},\\
Y(t)=C_{out} X,
\end{array}
\end{equation}
where $U$ represents the control input force on each motor and $G_{d}$ is the influence of ground effects. The fusion controller then triggers switching based on the height of the robot from the ground. 

This task was considered an infinite horizon optimal problem and employed an LQR-based controller to stabilize the robot's attitudes. We introduce the slack variable $s$ to transform the robot's states inequality constraint into an equality as follows:
\begin{equation}
\begin{array}{l}
min (J_{\Sigma}) =\frac{1}{2} \int_{t_{0}}^{\infty }\left(Y^{T} Q Y + U^{T} R U  +  G^{T} T G +  s^{T} P s \right) d t,
\end{array}
\end{equation}
subject to
\begin{equation}
\begin{array}{l}
Y_{\min }-s \leq Y \leq Y_{\max }+s, \\
G_{\min }-s \leq G \leq G_{\max }+s, \\
\Delta U_{\min } \leq \Delta U \leq \Delta U_{\max }, \\
0 \leq U \leq U_{\max },\\
s \geq 0.
\end{array}
\end{equation}
A slack variable $s$ was then introduced to solve the cost function in a larger feasible region. In this process, the robot solves for an optimal motion trajectory using Eqs. (\ref{cost}-\ref{expect}), as an auxiliary controller assists in adjusting the posture before landing. The pitch angle then determines whether landing on a slope can be performed safely. The potential for smooth mode-switching from flight to driving is then dictated by whether the yaw angle and the speed direction remain consistent.

\section{Validation and Discussion}

This section describes validation experiments consisting of robot simulations. A detailed model was constructed in ADAMS using equations discussed in Section III and parameters provided in Table \ref{specification}. The fusion controller described in Section IV was also evaluated. For comparison purposes, a parallel cascade PID controller was independently employed during flight, ground travel, and structural deformation to provide a baseline. This controller generated rotational speed in each motor (for flight control and chassis steering) and wheel speed (for ground motion).

\subsection{Vibration Tests in Ground Driving}
A foldable rotary mechanism can reduce damage to the arms caused by vibrations. In the simulations, we tested the effects of vibrations with and without the deformable structures. Simulated slope landing also verified the robustness and validity of the proposed fusion controller.

\subsubsection{Simulation Tests}

\begin{figure}[b]
    \centering
    \includegraphics[width=6cm]{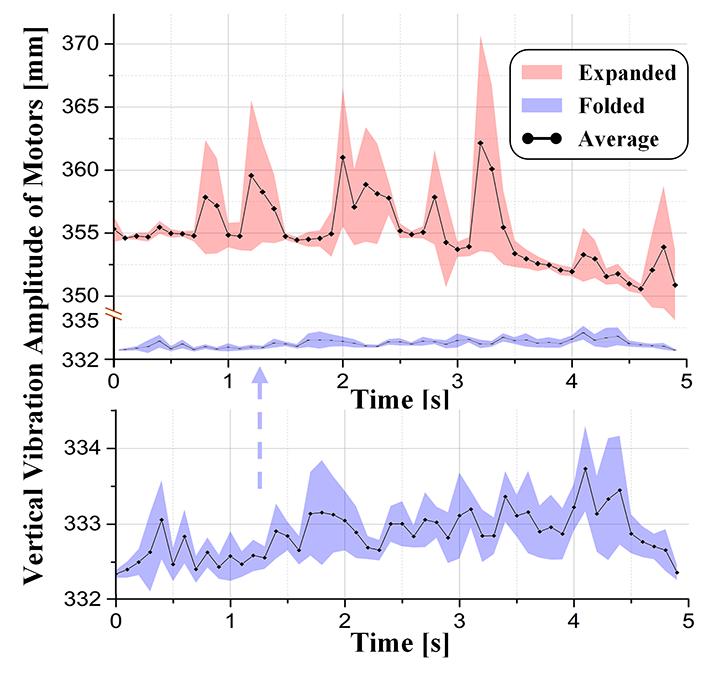}
    \caption{Vibration amplitude in the folded and unfolded states in simulations.}
    \label{tolerance}
\end{figure}

In addition to reducing the volume of the robot for passing through narrow regions, the deformable mechanism can also reduce vibrations caused by fit tolerances. Coupled with the movable structure and scattered robot mass distribution, the significant vibrations generated when driving on the ground can cause irreversible damage to these mechanisms. Vibration damping performance, before and after deformation, was assessed using the average vertical displacement $ H $ of the four motors as a reference. Fig. \ref{tolerance} displays error diagrams for $ H $ in the unfolded and folded states. As the robot arm unfolds during ground driving, severe vibrations will be generated and transmitted to the motor, an effect caused by the outward extension of the arm. This deformable mechanism exhibits a high fit tolerance and the extended mass dynamic system reached a maximum shaking displacement of 25 mm. The vibrations generated during driving are then significantly reduced when the arm is folded, approaching a maximum of $\sim$3 mm.

\begin{figure}[t]
    \centering
    \includegraphics[width=7cm]{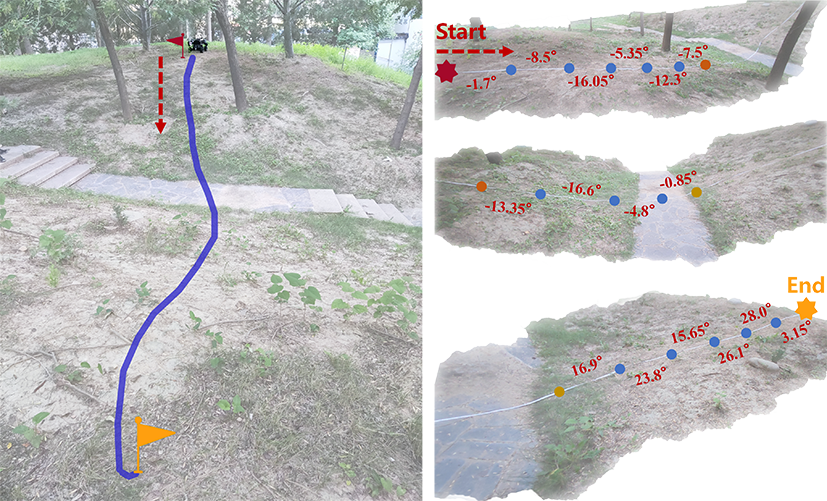}
    \caption{Unstructured scene details for ground motion experiments.}
    \label{field}
\end{figure}

\begin{figure}[t]
    \centering
    \includegraphics[width=7cm]{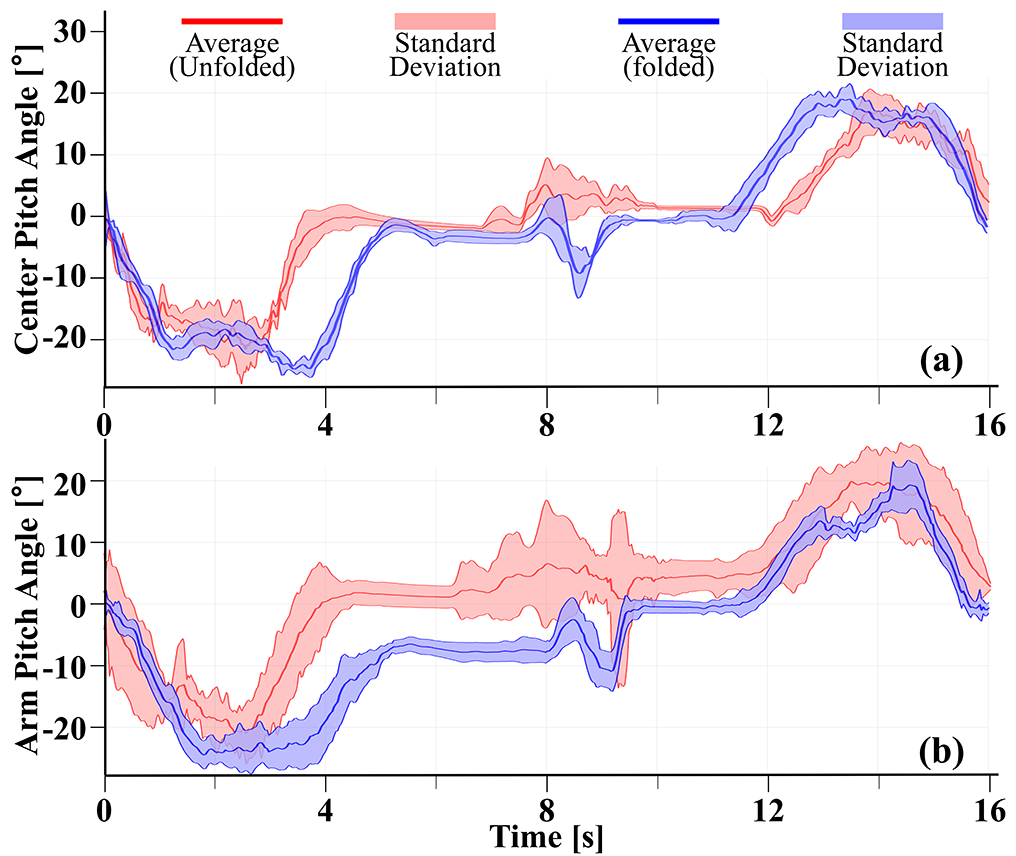}
    \caption{The average and standard deviation for angular transformations in different modes used to observe vibrations.}
    \label{driving experiment}
\end{figure}

\subsubsection{Field Experiments}
The robot was programmed to drive through a field with steep slopes approaching $28^\circ $, as shown in Fig. \ref{field}. The slope angles were recorded for each 1 m segment and used as a reference. Five inertial sensor modules (IMU) were mounted to the robot's center and the four motors. Vibrations in the robot, caused by fit tolerances, were evaluated by measuring angle changes at key positions (Center of gravity and four motors). Loss of control and ground slip during movement were also assessed \cite{liao2018model} by the trends and shifts in results. A global positioning system (GPS) was mounted to the robot to acquire location information, as the robot repeated 10 sets of driving experiments on a slope at a speed of 1 $m/s$. Both extended and folded modes were included in 5 sets of experimental data. Fig. \ref{driving experiment} shows fitting results for pose angle changes at each critical position in the two modes. Front arms, rear arms, and center average data are respectively indicated by lines, with the standard deviation shaded around the average in the figure. As shown, the robot maintained excellent ground maneuverability with the arms folded. In the experimental comparison of these two states, the unfolded arms caused greater amplitude and frequency vibrations, with vibrations of the center position similar among the two motion modes. With the arms unfolded, the robot speed was often out of control, slipping at 3-5 seconds and 11-13 seconds. As such, trajectory offsets are evident and the robot is more difficult to control, steer in the unfolding mode than during ground driving.

\subsection{Control Tests for Ramp Mode Switching}
\begin{figure}[h]
    \centering
    \includegraphics[width=8.6cm]{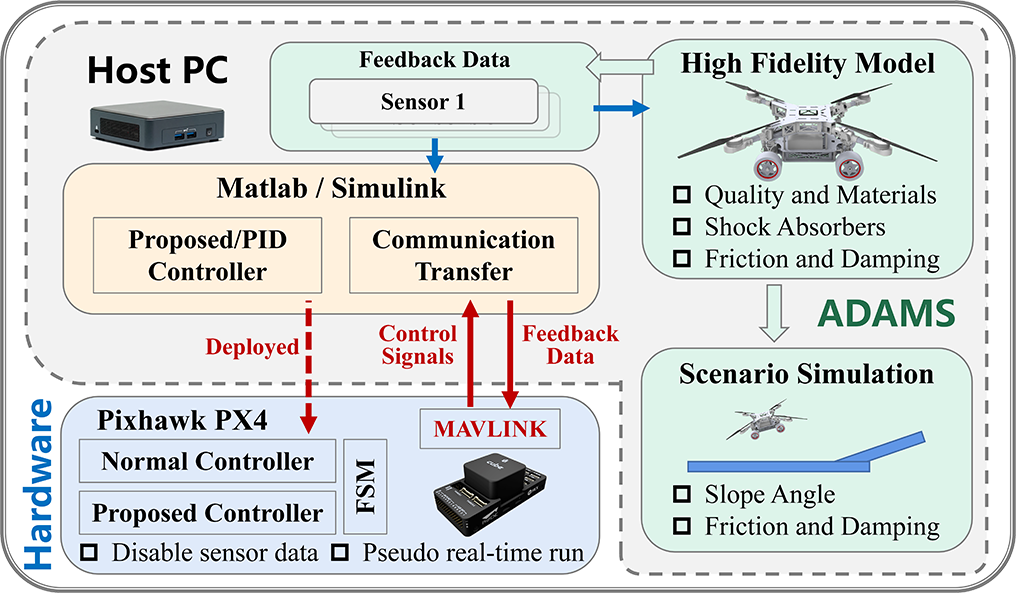}
    \caption{The hardware-in-the-loop test platform. Control code was generated using Simulink and deployed on the PX4 autopilot hardware. A high-fidelity model constructed in ADAMS communicated with the PX4 through Simulink.}
    \label{HIL}
\end{figure}

\begin{figure}[t]
    \centering
    \includegraphics[width=8.6cm]{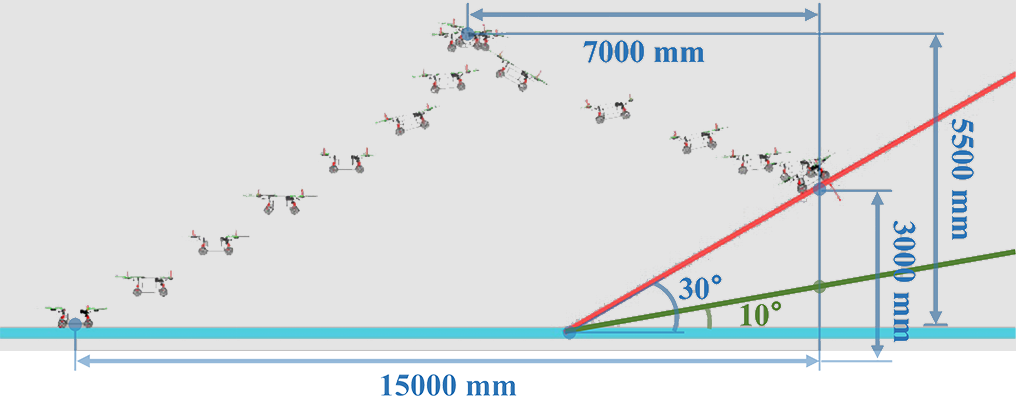}
    \caption{The simulation environment and motion trajectories for 10° and 30° scenes.}
    \label{track_10}
\end{figure}

\begin{figure}[h]
    \centering
    \includegraphics[width=7cm]{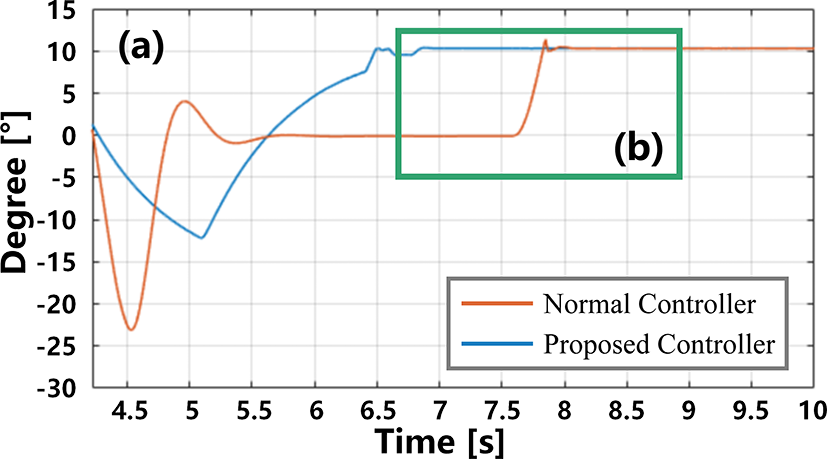}
    \caption{Trends in the robot pitch angle for a 10° scene. (a) Variations in the pitch angle from hovering to landing. (b) The landing process.}
    \label{10pitch}
\end{figure}

\begin{figure}[h]
    \centering
    \includegraphics[width=7cm]{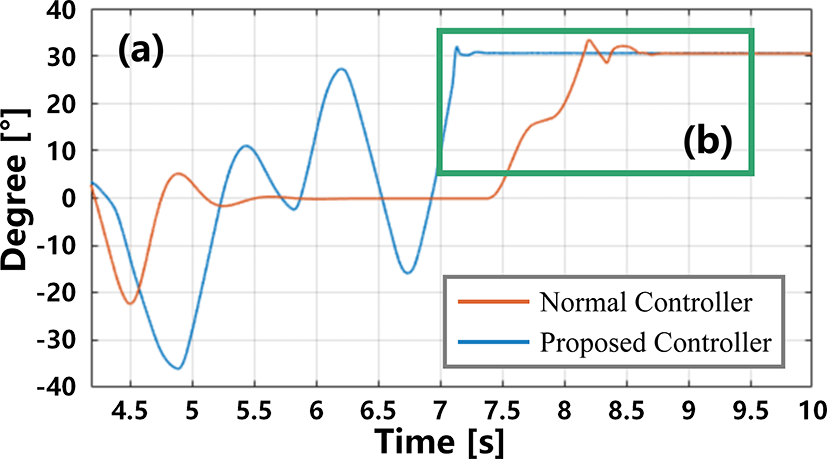}
    \caption{Trends in the robot pitch angle for a 30° scene. (a) Variations in the pitch angle from hovering to landing. (b) The landing process.}
    \label{30pitch}
\end{figure}

\begin{figure}[t]
    \centering
    \includegraphics[width=7cm]{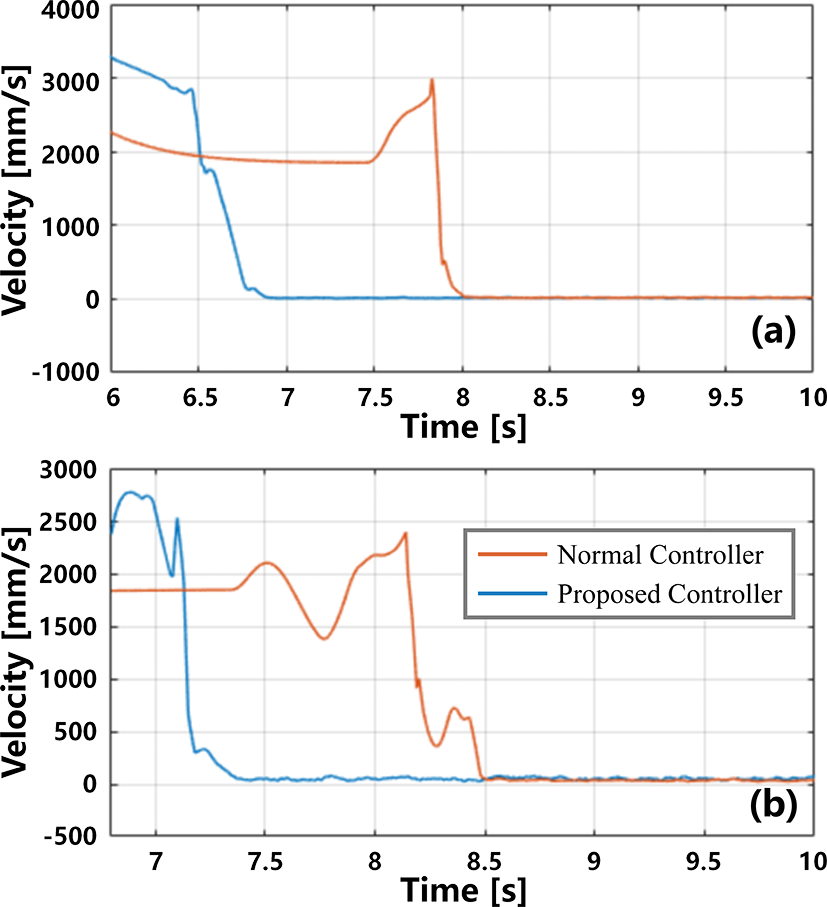}
    \caption{Variations in the robot's velocity for (a) 10° and (b) 30° scenes.}
    \label{velocity}
\end{figure}

\begin{figure}[h]
    \centering
    \includegraphics[width=7cm]{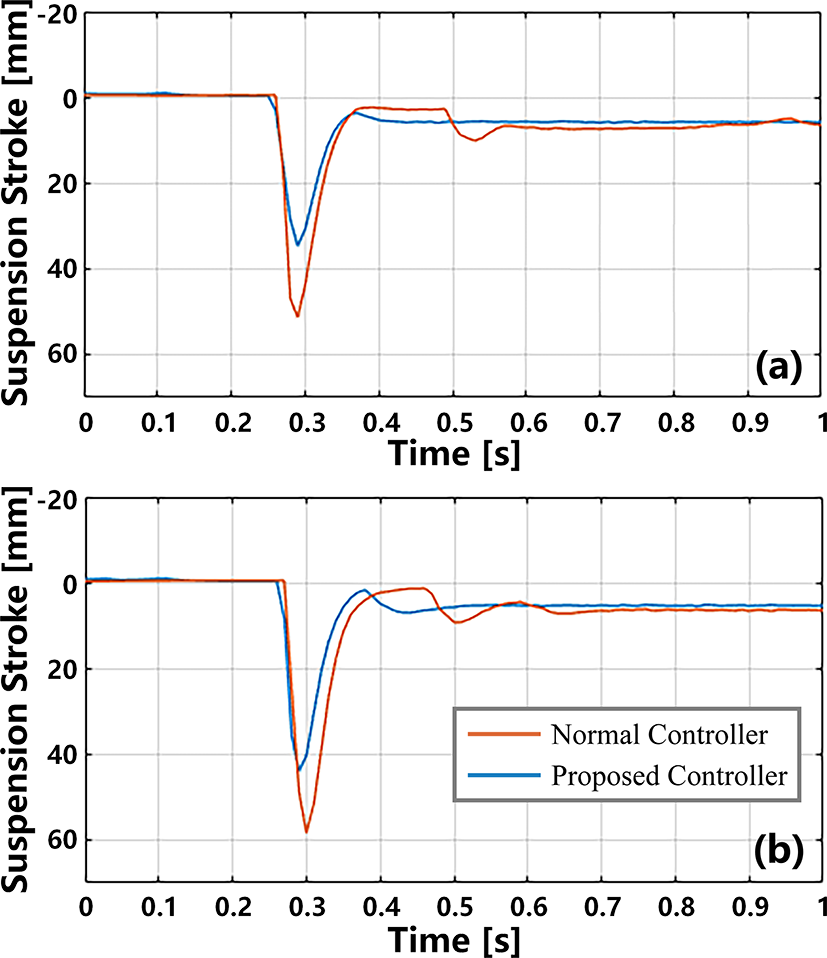}
    \caption{Variations in the robot's suspension stroke for (a) 10° and (b) 30° scenes.}
    \label{Springer}
\end{figure}

Mode switching on slopes requires motion planning and adjusting both speed and attitude before landing. Slopes are typically no more than 10 degrees for general urban or roadway scenes. Furthermore, for rotor-driven robots, a 30-degree slope is a challenging scenario for both takeoff and landing. If the robot could perform flexible and compliant mode switching in these two scenarios, it could also perform tasks in unstructured environments such as earthquake ruins and city fire scenes. As such, a hardware-in-the-loop (HIL) platform was developed to accommodate the high-fidelity robot model and the PX4 autopilot hardware. HIL experiments, which have been shown to be effective for testing embedded control systems, were used for performance evaluation. The robot model was established in ADAMS and Matlab/Simulink using detailed parameters (Tab.\ref{specification}), as shown in Fig. \ref{HIL}. A normal controller with parallel cascade PID modules was included for comparison, to verify the robustness and effectiveness of the proposed controller. Verification experiments were conducted on 10° and 30° slopes as shown in Figs. \ref{track_10}-\ref{Springer}. Optimization of the proposed controller used for slope mode switching was conducted by characterizing the pitch angle, velocity, and spring travel, as shown in Figs. \ref{10pitch}-\ref{velocity}.

Trends in pitch angle during various landing tasks are shown in Figs. \ref{10pitch} and \ref{30pitch}. The proposed controller could adjust the robot's pitch angle to be consistent with the slope prior to landing. In this process, the robot will not fly backwards or oscillate by leaning backward. The section of the graph in the green rectangle shows oscillations in the pitch angle during landing, an indication of whether the landing process was smooth. As the robot with the proposed onboard controller landed in the 10° scene, it became stable almost instantly. However, the robot with the normal controller exhibited a pitch angle of 2° with induced oscillations. In the 30° scene, the proposed controller was quickly stable within 0.2 s after landing, with a maximum pitch angle of $\sim$2.5°. The maximum oscillation angle for the normal controller exceeded 4° and tended to stabilize after two oscillations, $\sim$0.5 s after landing. Fig. \ref{velocity} shows velocity curves perpendicular to the direction of the descent slope. Similar to the pitch angle, these data characterize smoothness during landing. Finally, Fig. \ref{Springer} shows suspension strokes reflecting the magnitude of the impact force when landing without optimized control. In the 10° scene, the suspension stroke for the robot with the proposed controller was $\sim$34.3\% less than that of the normal controller. In addition, suspension convergence times were 0.12 s and 0.31 s, respectively. In the 30° scene, the suspension stroke for the robot with the proposed controller was $\sim$21.7\% less than with the normal controller. The corresponding suspension times were 0.28 s and 0.44 s, respectively.

\begin{table*}[!t]
\renewcommand\arraystretch{1.1}
\begin{center}
\caption{Undisturbed landing times offset from the robot landing point with random disturbances.}
\label{land}
\begin{tabular}{cccccc}
\hline \text { \textbf{Controller} } & \text { \textbf{Landing Time} } & \text { \textbf{Offset (40 N)} } & \text { \textbf{Offset (60 N)} } & \text { \textbf{Offset (80 N)} } & \text { \textbf{Mean Variance} } \\
\hline \text { Proposed Controller }$\left(10^{\circ}\right)$ & 3.30 {~s} & 282.99 {~mm} & 343.60 {~mm} & 243.15 {~mm} & 292.84 {~mm} \\
\text { Normal Controller }$\left(10^{\circ}\right)$ & 3.77 {~s} & 468.76 {~mm} & 500.81 {~mm} & 576.72 {~mm} & 517.41 {~mm} \\
\text { Proposed Controller }$\left(30^{\circ}\right)$ & 3.00 {~s} & 229.30 {~mm} & 220.00 {~mm} & 372.27 {~mm} & 282.59 {~mm} \\
\text { Normal Controller }$\left(30^{\circ}\right)$ & 3.98 {~s} & 391.40 {~mm} & 339.26 {~mm} & 488.07 {~mm} & 410.89 {~mm} \\
\hline
\end{tabular}
\end{center}
\end{table*}

We also measured the robot's landing speed when perturbed only by ground effects. In the 10° slope descent task, the normal and proposed controllers required 3.77 s and 3.3 s to land, respectively, representing a 14.2\% increase in speed. In the 30° slope landing task, the normal controller required 3.98 s to land and the proposed controller required 3 s, a speed increase of 32.7\% with the proposed model. Controller robustness was verified by adding forces of 0-40 N, 60 N, and 80 N at the robot center of mass to simulate external disturbances, the directions and magnitudes of which were randomly generated. In this setting, the robot performed 10 experiments with different controllers in each of the two scenarios. Landing accuracy was characterized using the mean variance of the landing offsets between the actual landing point and the expected landing point, as shown in Table \ref{land}. Under random disturbances, the proposed controller achieved notable improvements in speed and landing accuracy compared with the normal controller.

\section{Conclusion}

This paper introduced a deformable land-air robot with continuous switching capabilities, including the establishment of a detailed coupled dynamics model. The design of the deformable arms, locking mechanism and limit motors enable the robot to move flexibly and stably both in flight and driving. A fusion controller based on JLT and LQR was designed to focus on unstructured scenes and land-air switching motion. The controller considered control assignments, land-air actuators, and ground-effect disturbance, which enabled the robot to achieve fast and robust mode switching. Combined with a coupled dynamics model and the proposed fusion controller, a hardware-in-the-loop system was implemented for a high-fidelity robot model based on ADAMS/Simulink. Ground driving and mode-switching experiments were conducted using the HIL system and in practice. Results showed these deformable structures effectively improved robot ground-driving accuracy. Compared with the normal controller, the proposed controller significantly improved the results of trajectory planning and attitude controling. The speed of landing  and mode-switching flexibility in sloped scenes. Compared with the PID controller, the proposed controller has a maximum increase of about 24.6 \% in landing speed. On this basis, the generated trajectory and the auxiliary controller reduce the robot's offset of landing point ~43.4 \%. The improvement in flexibility and compliance is reflected in the impact on the robot is reduced by ~34.3 \%. The convergence speed of posture and obstacle avoidance after landing also doubles.

\section{ACKNOWLEDGMENTS}

This work was supported by the National High Technology Research and Development Program of China under Grant No. 2018YFE0204300 and the National Natural Science Foundation of China under Grants No. 62273198 and U1964203. We thank LetPub (www.letpub.com) for linguistic assistance and pre-submission expert review.

\bibliographystyle{IEEEtran}
\bibliography{0article}

\vspace{-25 mm} 


\begin{IEEEbiographynophoto}{}

\begin{IEEEbiography}
[{\includegraphics[width=1in,clip,keepaspectratio]{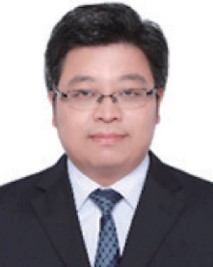}}]{Xinyu Zhang} was born in Huining, Gansu Province, and he received a B.E. degree from the School of Vehicle and Mobility at Tsinghua University, in 2001. He was a visiting scholar at the University of Cambridge.
\newline He is currently a researcher with the School of Vehicle and Mobility, and the head of the Mengshi Intelligent Vehicle Team at Tsinghua University. Dr. Zhang is the author of more than 30 SCI/EI articles. His research interests include intelligent driving and multimodal information fusion.
\end{IEEEbiography}

\vspace{-10 mm} 

\begin{IEEEbiography}[{\includegraphics[width=1in,clip,keepaspectratio]{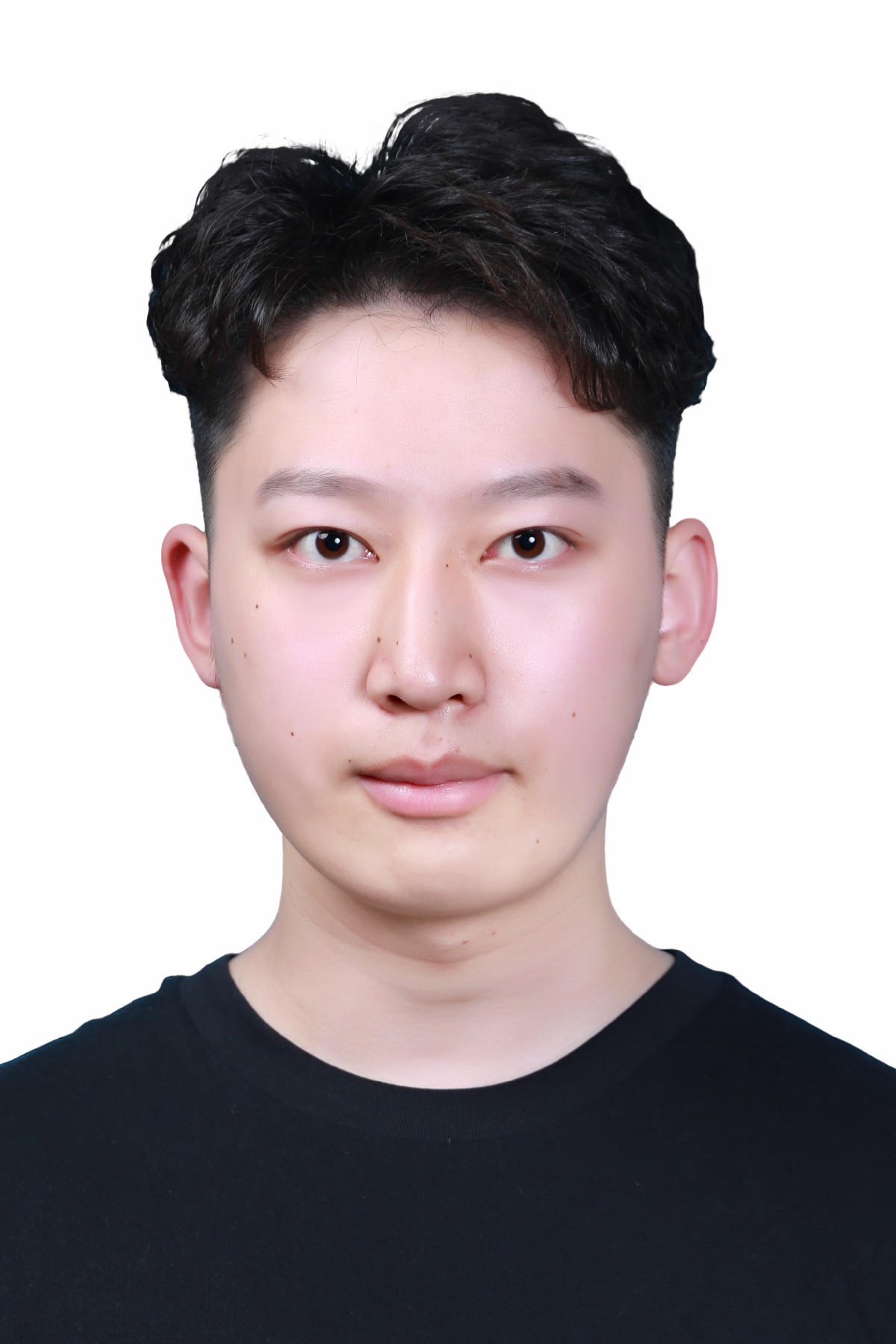}}]
{Yuanhao Huang} was born in Chengdu, Sichuan Province, and he received the B.E. degree from the Institute of Disaster Prevention Department, Beijing, China, in 2020. Now he is pursuing a Master degree in engineering at Inner Mongolia University of Technology, Hohhot, China. 
He is currently working on a joint training program at New Technology Concept Automobile Research Institute, Tsinghua University. His research interests include robotics, automatic system and motion planning.
\end{IEEEbiography}

\vspace{-10 mm} 

\begin{IEEEbiography}
[{\includegraphics[width=1in,clip,keepaspectratio]{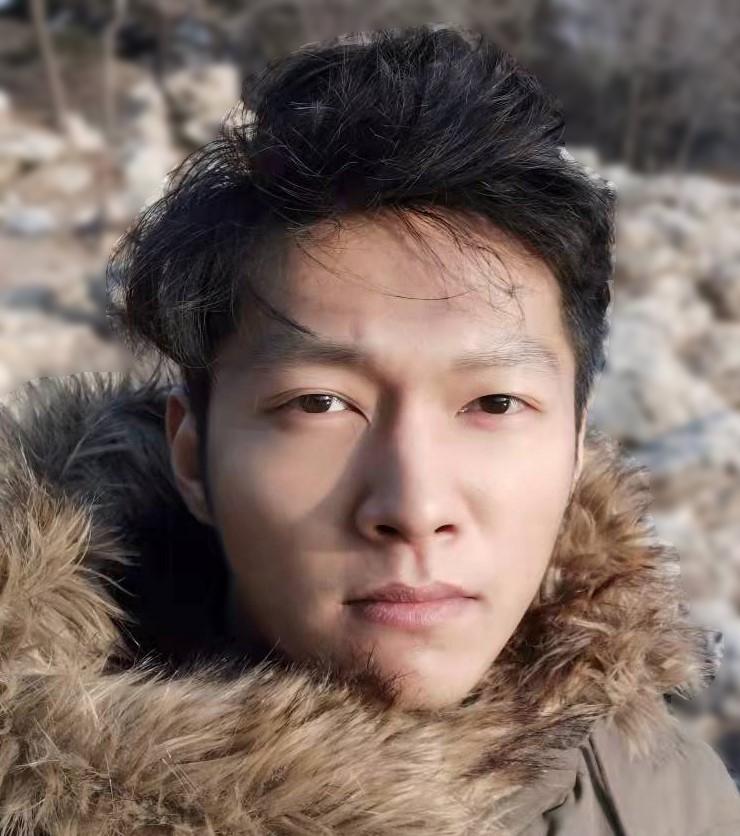}}]{Kangyao Huang} received the B.Eng. degree in Aerospace from Northwestern Polytechnical University, Xi’an, China, in 2016, and the M.Res. degree in Control \& Systems Engineering from the University of Sheffield, Sheffield, U.K., in 2020. Currently he is pursuing a Ph.D degree at the Department of Computer Science and Technology, Tsinghua University, Beijing, China, and working with the Mengshi Intelligent Vehicle Team.
\newline He has three years working experience in aerospace industry. He was the Founder, Technical Director and Chief Engineer with Bingo Intelligence Aviation Technology co., LTD, where he developed the general software architecture for integrated avionics system, and provided applied research in cooperation with partners in aerospace and manufacturing sectors. His research interests include swarm robotics, UAV engineering, and embodied intelligence.
\end{IEEEbiography}

\vspace{-10 mm} 

\begin{IEEEbiography}
[{\includegraphics[width=1in,clip,keepaspectratio]{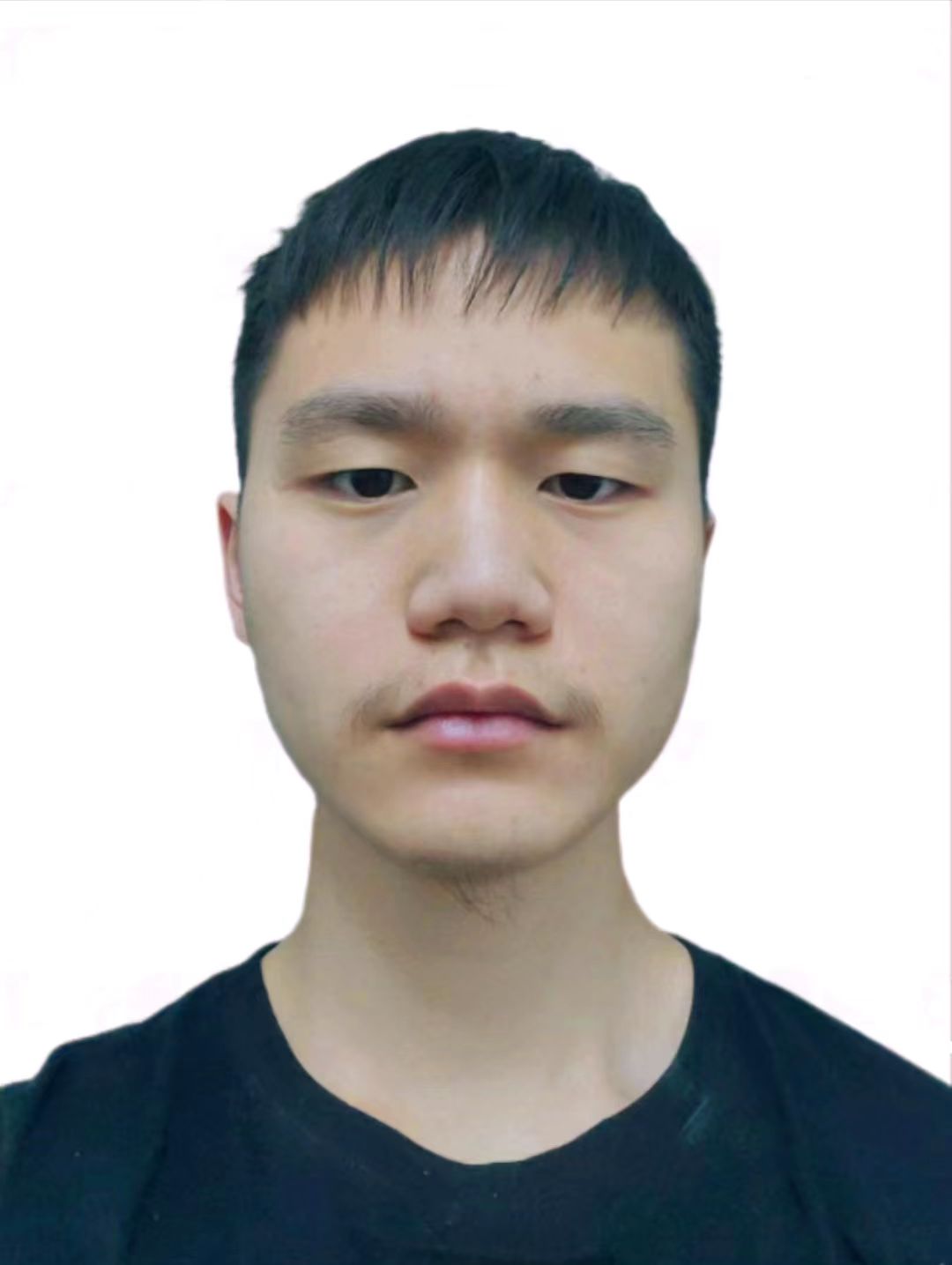}}]{Ziqi Zhao} was born in Tongren, Guizhou province. Now he is studying for a Bachelor's degree in Engineering at the Institute of Disaster Prevention Science and Technology. He is currently doing a research internship at Tsinghua University. His research interests include robotics, automatic control and SLAM and Navigation2.
\end{IEEEbiography}

 \vspace{-10 mm} 
 
 \begin{IEEEbiography}
[{\includegraphics[width=1in,clip,keepaspectratio]{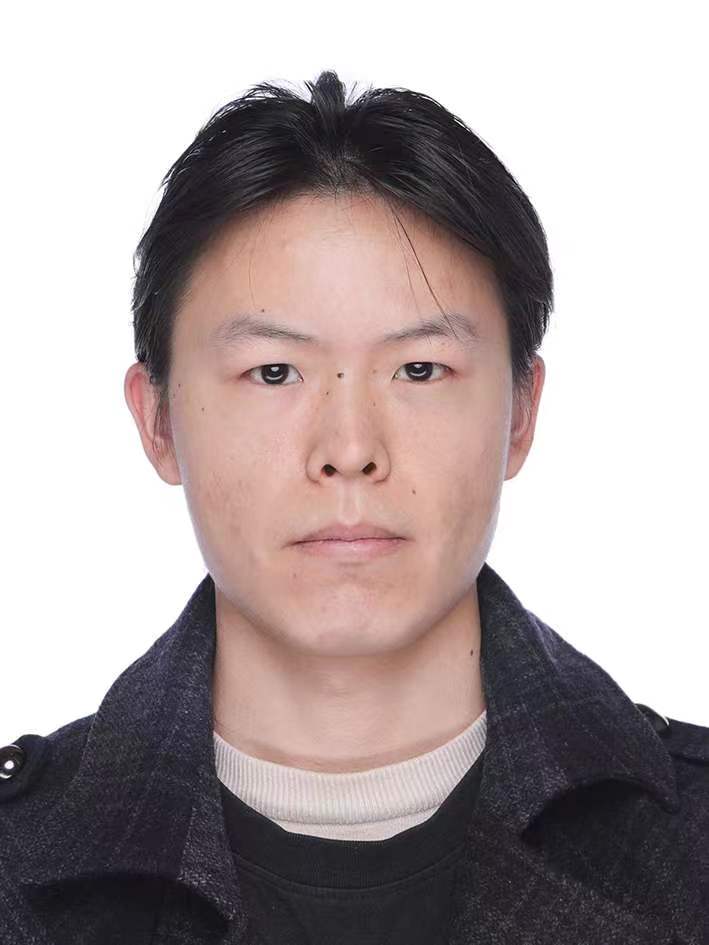}}]{Jingwei Li} was born in Jinghe,Xinjiang Province,and he received the B.E. degree from the College of Mechanical and Electronic Control Engineering at the Beijing Jiaotong University,in 2021.Now he is pursuing a Master degree in engineering at Beijing University of Aeronautics and Astronautics.He has assisted in completing the simulation task of the Institute.His research interests include automatic control and analog simulation.
\end{IEEEbiography}

 \vspace{-10 mm} 
 
\begin{IEEEbiography}
[{\includegraphics[width=1in,clip,keepaspectratio]{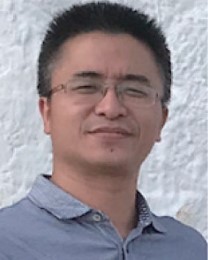}}]{Huaping Liu} (Senior Member, IEEE) is an Associate Professor with the Department of Computer Science and Technology, Tsinghua University, Beijing, China. His research interests include robot perception and learning. Dr. Liu has served as an Associate Editor of ICRA and IROS and in the Program Committees of IJCAI, RSS, and IJCNN. He is an Associate Editor of the IEEE ROBOTICS AND AUTOMATION LETTERS, Neurocomputing, and Cognitive Computation.
\end{IEEEbiography}

 \vspace{-10 mm} 

\begin{IEEEbiography}
[{\includegraphics[width=1in,clip,keepaspectratio]{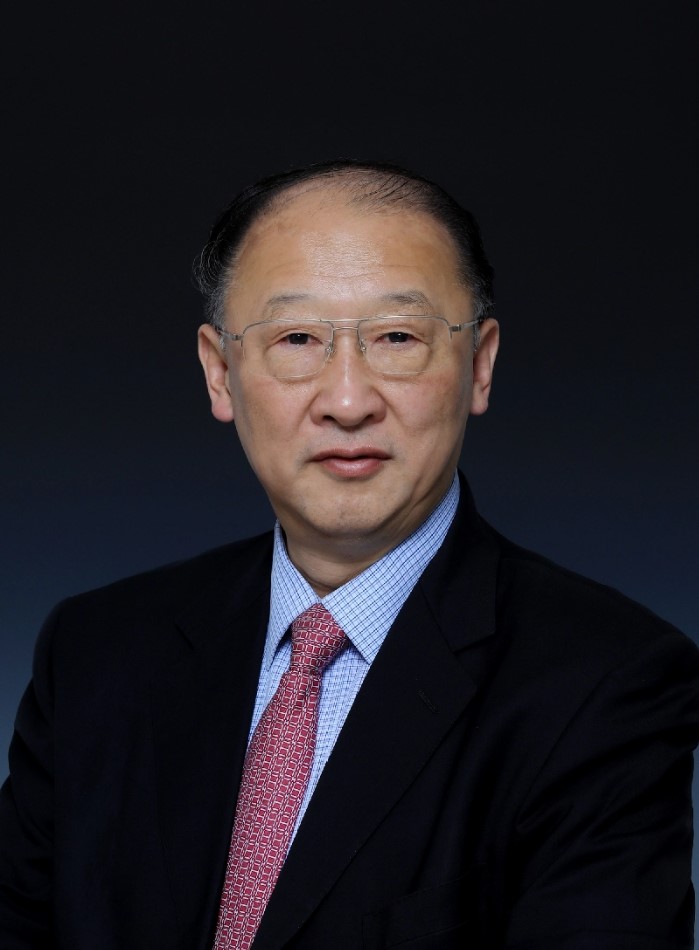}}]{Jun Li} (Fellow, Chinese Academy of Engineering) received the Ph.D. degree in internal combustion engineering from Jilin Polytechnic University, Changchun, China, in 1989. He is a fellow of the Chinese Academy of Engineering, and the Vice-Chief Engineer and the Director of the Research and Development Center with China FAW Group Corporation, Changchun.
\end{IEEEbiography}

\end{IEEEbiographynophoto}

\end{document}